\documentclass[sigconf]{acmart}

\AtBeginDocument{%
  }
\usepackage{subfig}

\usepackage{pifont}
\begin{document}


\title{Privileged Contrastive Pretraining\\for Multimodal Affect Modelling}


\author{Kosmas Pinitas}
\affiliation{%
  \institution{Institute of Digital Games\\ University of Malta}
  \city{Msida}
  \country{Malta}}
\email{kosmas.pinitas@um.edu.mt}
\orcid{0000-0003-0938-682X}

\author{Konstantinos Makantasis}
\affiliation{%
  \institution{Department of Artificial Intelligence\\ University of Malta}
  \city{Msida}
  \country{Malta}}
\email{konstantinos.makantasis@um.edu.mt}
\orcid{0000-0002-0889-2766}

\author{Georgios N. Yannakakis}
\affiliation{%
  \institution{Institute of Digital Games\\ University of Malta}
  \city{Msida}
  \country{Malta}}
\email{georgios.yannakakis@um.edu.mt}
\orcid{0000-0001-7793-1450}

\renewcommand{\shortauthors}{Pinitas et al.}


\begin{abstract} 
  Affective Computing (AC) has made significant progress with the advent of deep learning, yet a persistent challenge remains: the reliable transfer of affective models from controlled laboratory settings (\emph{in-vitro}) to uncontrolled real-world environments (\emph{in-vivo}). To address this challenge we introduce the \textit{Privileged Contrastive Pretraining} (PriCon) framework according to which models are first pretrained via supervised contrastive learning (SCL) and then act as teacher models within a Learning Using Privileged Information (LUPI) framework. PriCon both leverages privileged information during training and enhances the robustness of derived affect models via SCL. 
  Experiments conducted on two benchmark affective corpora, RECOLA and AGAIN, demonstrate that models trained using PriCon consistently outperform LUPI and end to end models. Remarkably, in many cases, PriCon models achieve performance comparable to models trained with access to all modalities during both training and testing. The findings underscore the potential of PriCon as a paradigm towards further bridging the gap between \emph{in-vitro} and \emph{in-vivo} affective modelling, offering a scalable and practical solution for real-world applications.  
\end{abstract}


\settopmatter{printacmref=false} 
\renewcommand\footnotetextcopyrightpermission[1]{} 

\begin{teaserfigure}
\vspace{-1em}
\centering
\includegraphics[width=0.5\textwidth]{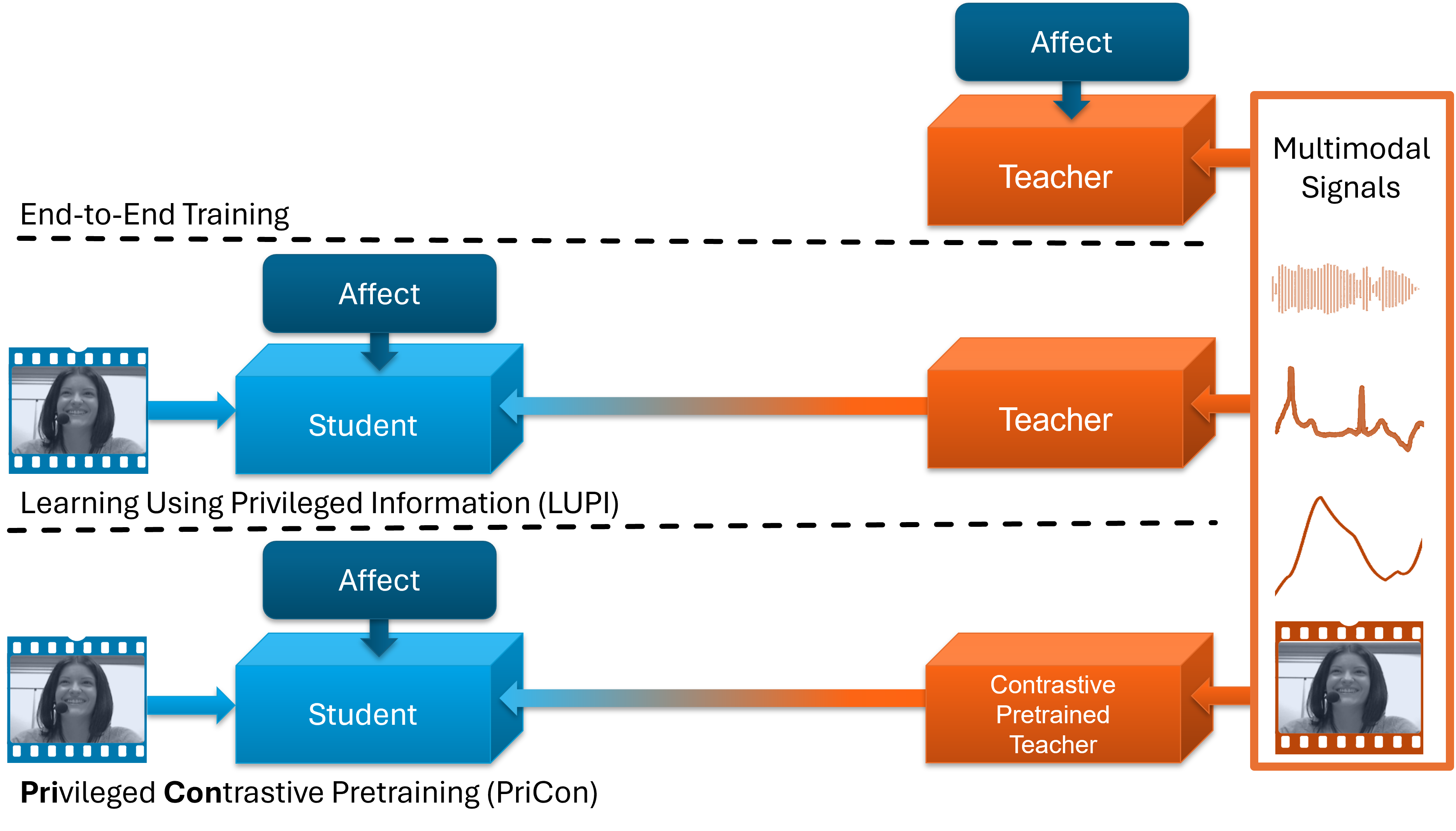}
\vspace{-1em}
	\caption{Visualising the proposed Privileged Contrastive Pretraining (PriCon) framework. The top figure illustrates the traditional end-to-end training of an affect model based on a set multimodal signals. The middle figure illustrates a LUPI paradigm for affect modelling, by which the student model learns affect labels from modalities available in the wild (e.g., visual frames in this example) while the teacher model uses additional multimodal signals available in the lab (e.g., audio and physiological signals in this example) during training. The bottom figure visualises the high-level concept of PriCon that trains a teacher model via SCL prior to transferring knowledge to the student via LUPI.} 
	\label{fig:teaser}
\end{teaserfigure}


\maketitle
\section{Introduction}



Affective Computing (AC) has made significant leaps forward in recent years due to advances in deep learning \cite{toisoul_estimation_2021,botelho_improving_2017,li_how_2020,trigeorgis_adieu_2016,tzirakis_end--end_2021,makantasis2021pixels}, which directly enhanced the predictive capacity of affect models. A major AC challenge remains, however: transferring models trained in controlled (\emph{in-vitro}) environments to real-world (\emph{in-vivo}) settings. Although laboratory conditions enable precise data collection, real-world scenarios introduce noise, privacy concerns, and hardware constraints, widening the \emph{in-vitro} to \emph{in-vivo} gap.

As a response to the above challenges this paper introduces a unified framework that combines the advantages of Supervised Contrastive Learning (SCL) and the Learning Using Privileged Information (LUPI) paradigm \cite{vapnik2009new, vapnik2015learning}. The introduced \textit{Privileged Contrastive Pretraining} (PriCon) framework both exploits additional modalities during training as privileged information for the model and improves the generalisability of derived affect models deployed in \emph{in-vivo} scenarios (see Figure \ref{fig:teaser}). While LUPI allows PriCon to leverage richer data during training than what is available at test time, SCL enhances the quality of learned representations by encouraging class-wise separation in the feature space. We integrate these two components within PriCon to address two key hypotheses. First, LUPI can enhance affect detection under real-world (\textit{in-vivo}) conditions, producing performances comparable to \textit{in-vitro} models (\textbf{H1)}. Second, PriCon further enhances the effectiveness of \emph{in-vivo} LUPI models (\textbf{H2}). Both hypotheses are evaluated on the RECOLA \cite{ringeval2013introducing} and AGAIN \cite{melhart2021again} datasets, across different affective dimensions and modalities.


For testing H1, we incorporate fine-grained image features and fused data (i.e., image frames and features) as privileged information during \emph{in-vitro} training, while using only frames for \emph{in-vivo} testing. This approach reflects real-world limitations, where additional modalities (e.g., physiological signals) are impractical due to privacy and hardware constraints. For testing H2, PriCon is applied to \emph{in-vitro} models, refining their representations and improving their transferability to \emph{in-vivo} settings. Our results show that privileged information significantly enhances model performance (validating H1), while training privileged models via PriCon further improves \emph{in-vivo} performance (validating H2). Notably, PriCon trained models achieve performances comparable to those obtained from models trained on all modalities.


This paper makes several key contributions towards advancing affect modelling in real-world settings. First, we evaluate the LUPI paradigm for affect modelling in games and dyadic interactions, demonstrating its potential in dynamic environments. Second, we introduce the PriCon framework that boosts the robustness and transferability of \emph{in-vitro} LUPI models via SCL. Third, we validate the approach introduced on both RECOLA and AGAIN datasets, showing its robustness across affective dimensions (arousal and valence) and tasks (games vs. dyadic interactions). These findings underscore the benefits of \emph{privileged contrastive pretraining} in creating more generalisable affective models for real-world applications.

\section{Related Work}

This section reviews related work on the two key learning paradigms featured in PriCon. Specifically, we survey learning using privileged information and contrastive learning under the lens of affect modelling.

\subsection{LUPI for Affect Modelling}

While both distillation and LUPI involve a teacher-student framework, they serve different purposes. Traditional knowledge distillation focuses on compressing knowledge by training a student model to mimic the outputs of a more complex teacher model. For instance, Sun et al. \cite{sun2020dynamic} proposed a novel technique for micro-expression detection that distils knowledge from a pre-trained deep teacher neural network to a shallow student neural network via a teacher-student correlative framework enabling the student to capture fine-grained details necessary for accurate micro-expression recognition. Similarly, Liu et al. \cite{liu2022social} developed a cross-modal consistency modelling-based knowledge distillation framework for image-text sentiment classification of social media data. By aligning the information between images and text, the student model can effectively capture the sentiment expressed in multimodal social media posts.

Several affective computing problems have asymmetric train and test input distributions. This discrepancy means that the training phase might have access to richer or more diverse data than what is available during testing. Hence, it is not surprising that the LUPI paradigm has started to become popular for AC research. LUPI addresses this asymmetry by allowing the use of additional information during training that is not available during testing. Makantasis et al. \cite{makantasis2021affranknet+} introduced a ranking model that treats additional training information as privileged information to rank affect states. This approach leverages richer data available during training, such as audio or contextual information, to improve the ranking model's accuracy in predicting affective states. In another study \citep{makantasis2021privileged}, the same authors predicted arousal from gameplay footage while treating telemetry and heart rate as privileged information. Using detailed physiological and telemetry data during training, the model could better capture the nuances of player arousal based solely on gameplay footage during testing. Zhang et al. \cite{zhang2021distilling} proposed a LUPI method to distil EEG representations via capsule-based architectures for both classification (positive, negarive, neutral emotion) and regression (vigilance) tasks. The capsule-based architectures captured spatial hierarchies in the data, making the distilled knowledge more effective for the student model. 

This work extends LUPI-based affect modelling by applying it to affect classification, a previously unexplored area. Unlike prior studies that focus on ranking affect states or knowledge distillation for efficiency, our approach leverages multimodal privileged information to improve the discriminative power of the unimodal student model, enabling more accurate classification of affective states. Additionally, we enhance model robustness by pretaining the teacher models via SCL, improving generalisation.
\vspace{-1em}
\subsection{Contrastive Learning for Affect Modelling}

Contrastive learning techniques are among the most widely applied methods for learning representation. Notably, Li et al. \cite{li2021contrastive} investigated the impact of unsupervised representation learning for speech emotion recognition. Their study demonstrated that the proposed contrastive predictive coding method based on InfoNCE produced representations that achieved state-of-the-art performance across the activation, valence, and dominance dimensions.

Building on these advances, Mai et al. \cite{mai2022hybrid} introduced a novel hybrid contrastive learning framework. This framework integrates intra-modal, inter-modal, and semi-contrastive learning to enable the model to explore cross-modal interactions and preserve inter-class relationships, thereby reducing the modality gap. Their approach demonstrated significant improvements in capturing the complex relationships between different modalities, such as audio and visual data, in emotion recognition tasks. Pinitas et al. \cite{pinitas2022supervised} employed supervised contrastive learning on fine-grained multimodal features to develop robust arousal-infused representations. Their work underscored the importance of incorporating supervised learning signals to enhance the quality of representations, leading to higher arousal classification accuracy than the end-to-end alternatives. Yang et al. \cite{yang2023cluster} proposed an innovative low-dimensional supervised cluster-level contrastive learning method. This approach reduces the high-dimensional SCL space to a more manageable three-dimensional affect representation. By efficiently compressing the feature space, their method facilitates more effective and computationally efficient emotion recognition while maintaining high performance. Recently, Pinitas et al. \cite{pinitas2025across} introduced a novel few-shot representation learning framework that fine-tuned self-supervised models for engagement state prediction across FPS games outperforming end-to-end baselines.

Unlike the aforementioned works that focus on unsupervised, hybrid, or cluster-level contrastive learning for affect modelling, this paper integrates supervised contrastive learning within the LUPI paradigm with the aim to improve in-vitro to in-vivo generalisation. Additionally, the focus on affect classification in games and dyadic interactions, rather than general emotion recognition, sets this study apart by tackling dynamic and interactive environments where real-time adaptability is crucial and very much needed.

\section{Methodology}
This section first describes the use of privileged information and supervised contrastive learning in a concise way and then moves on to outline the model architectures employed. 

\subsection{Learning Using Privileged Information}

Learning using privileged information \cite{vapnik2009new, vapnik2015learning} addresses problems characterised by an asymmetric distribution of information between training and test time. LUPI provides the means to \textit{transfer knowledge} from all available modalities to a machine learning model that makes predictions using only a subset of these modalities \cite{sharmanska2013learning,lopez2016unifying}. As far as the RECOLA database is concerned, we treat as privileged the information that corresponds to physiology and ausiovisual features provided by the database creators \cite{ringeval2013introducing}. For AGAIN, we consider fine-grained gameplay features as privileged information \cite{melhart2021again}. Our choice is justified by the fact that capturing physiology requires specialised sensors, while constructing physiology and audiovisual features implies the employment of specific software algorithms. In the same vein, information about the state of the game (e.g., number and actions of enemies) requires access to the game engine itself. On the contrary, information that comes from raw footage frames is considered prevalent information as it can be captured using conventional cameras that can be available at both training and test times. 

This study explores the use of privileged information with neural network-based affect models. Following \cite{hinton2015distilling,lopez2016unifying,vapnik2017knowledge}, we represent privileged knowledge—i.e., information available only during training, such as physiological signals or high-resolution feature embeddings—within the output of a neural network that has been trained to make predictions based on either all available modalities (privileged + standard) or on privileged information alone. This model is called \textit{teacher}. A trained teacher model can transfer knowledge obtained through privileged information to another model called \textit{student}. The transfer of knowledge can be achieved by feeding the model with only those modalities of information that are available in the wild and forcing it during training to balance (via hyperparameter $\alpha$) between learning the task (cross-entropy loss $L_{CE}$) and learning prediction distributions that match those of the teacher model (KL divergence loss $L_{KL}$): 

\begin{equation}
	L_{P} = (1-\alpha) L_{CE}(f(x),y) + \alpha L_{KL}(g(\tilde{x}),f(x))
	\label{eq:probability_transfer_lupi_loss}
\end{equation}
The resulting optimisation objective $L_p$ is shown in Eq. \eqref{eq:probability_transfer_lupi_loss}. Where $f(x)$ and $g(\tilde{x})$ are the outputs of the student and teacher models, respectively, while $\tilde{x}$ refers to the privileged modalities. It should be noted that after training, the student model makes predictions based only on the information that is available in the wild, without any dependence on the teacher model or privileged information.

\subsection{Privileged Contrastive Pretraining}

A central component of the PriCon framework is the pretraining of teacher models using SCL, which lays the foundation for effective privileged supervision. By leveraging class label information, SCL guides the formation of semantically meaningful sample pairs, resulting in highly discriminative representations. This results in an embedding space where similar instances are clustered together, and dissimilar ones are pushed apart, enhancing the model’s capacity to generalise.

In the PriCon framework, this teacher pretraining stage is not merely auxiliary but is instrumental in learning rich, class-consistent representation structure. Specifically, for each input \( x_i \) (referred to as the \textit{anchor}), SCL constructs a set of \textit{positives} \( P_i \), which includes all other instances in the batch sharing the same class label, and contrasts them against \textit{negatives}---instances from different classes. This structured approach enhances the teacher's ability to model fine-grained intra-class similarities and inter-class separability, which is later transferred to the student model through the LUPI paradigm.

Compared to traditional supervised learning, which typically optimises for loss minimisation without explicitly shaping the latent space, SCL provides a more expressive training signal. By encouraging consistent semantic alignment in the representation space, it equips the teacher with a more nuanced understanding of class distributions---an essential property for the downstream knowledge transfer in PriCon. Formally, the supervised contrastive loss \( L_{SC} \) is defined as:

\begin{equation}
	L_{SC} = \sum_{i\in I} \frac{-1}{|P_i|}\sum_{p \in P_i}\log\frac{\exp(r_i\cdot r_p/\tau)}{\sum_{a\in A_i}\exp(r_i\cdot r_a/\tau)}
	\label{eq:scl_loss}
\end{equation}

\noindent where $I$ is a set that includes all samples and $P_i$ is the set that includes only the samples that are assigned to the same class as $i$. $A_i$ is a set that contains any element of set $I$ besides element $i$. With $r_i$, $r_p$ and $r_a$ we denote the latent representations of the model for the samples $i$, $p$ and $a$, respectively. Finally, $\tau$ stands for a non-negative temperature hyperparameter that controls the sharpness of the similarity distribution. It should be noted that $i$, $p$ and $a$ correspond to the index of the current sample, a sample positive to the current sample, and a sample different from the current one, respectively. 

This formulation ensures that representations of samples from the same class are tightly clustered by maximising \( r_i \cdot r_p \), while dissimilar samples are discouraged from occupying nearby regions in the latent space via the denominator term. In the context of PriCon, SCL serves as a powerful mechanism to encode privileged information into the teacher model, thus laying a strong representational foundation for effective student training in the subsequent knowledge transfer phase.

\subsection{Model Architectures}

\begin{figure}[!tb]
	\centering	\includegraphics[width=\columnwidth]{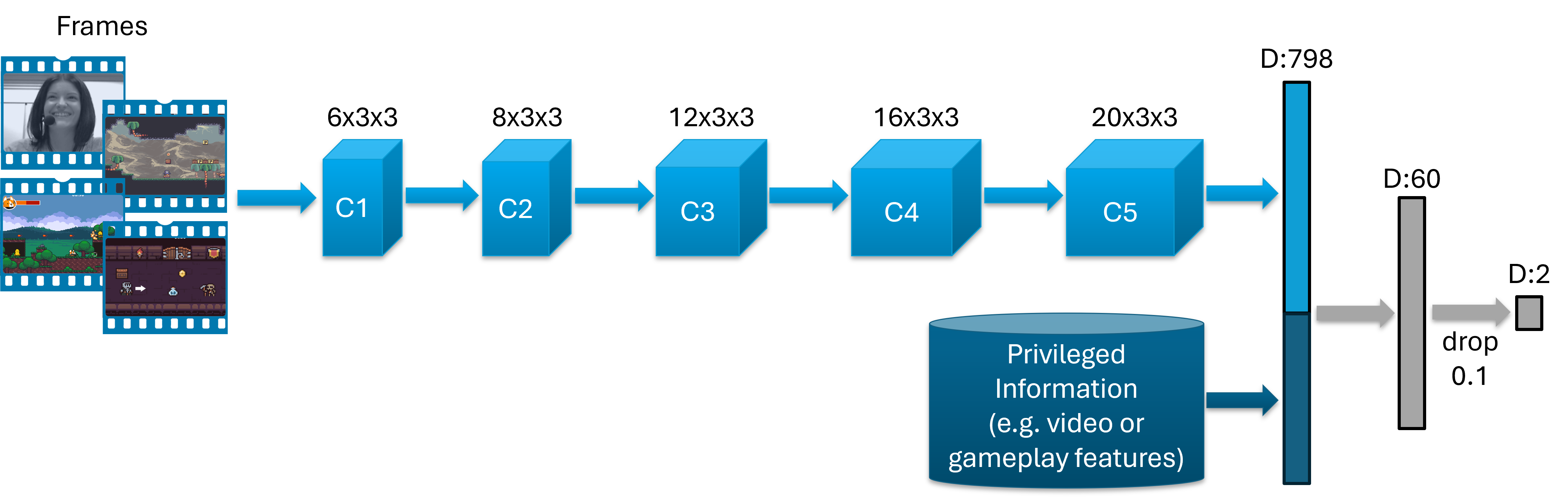}
    \vspace{-1em}
	\caption{Illustration of the affect models. `C' and `D' denote convolutional and dense layers, respectively. The student model (light blue) and privileged teacher (dark blue) are fused in the fusion teacher via the gray module.} 
	\label{fig:lupi_ch_teacher_and_student}
    \vspace{-1em}
\end{figure}

In this section, we first present the main components used in LUPI, namely student and teacher models. Then, we present the baseline architectures that we compare against our method for assessing the effectiveness of the obtained models. Both student and teacher architectures are illustrated in Figure \ref{fig:lupi_ch_teacher_and_student}.


\subsubsection{Student Model}\label{sec:lupi_ch_methodology_models_studet}
The student neural network $S$ is a predictive model that learns to estimate the target outcomes by utilising the information from frames. During the training phase, the student model leverages knowledge distilled from a teacher model $T$, which has access to an additional set of modalities (i.e., privileged information). This privileged information, available only during training, enhances the learning process by providing auxiliary guidance. The goal of the student model is to effectively generalise to new, unseen data, where privileged information is not available, using only the primary feature set. In this work $S$ is a Convolutional Neural Network (CNN) comprising five convolutional layers of 6, 8, 12, 16 and 20 filters, respectively, followed by a dense layer. The first four convolutional layers are configured with a stride parameter of 2, while the fifth convolutional layer uses a stride of 1. Each convolutional layer has a kernel size of dimensions 3x3 and employs the ReLU activation function. Finally, a dense layer with 768 neurons, also activated using ReLU, is applied to produce the encoded frame representation. The output of this layer is fed to the decision layer, which is a simple 2-neuron dense SoftMax-activated layer. A dropout of $0.1$ is applied before the last layer. 

\subsubsection{Teacher Models}\label{sec:lupi_ch_methodology_models_teachers}

The teacher, $T$, is a predictive model that is trained with access to both the primary feature set (i.e., prevalent information) and an additional set of privileged information. The teacher model is typically designed to achieve high performance by leveraging this richer information. Once trained, the teacher model's knowledge is distilled and transferred to the student model $S$, guiding it to improve its predictions based on the primary feature set alone. The teacher model plays a crucial role in the LUPI framework by acting as a source of enhanced supervision during training. In this paper, we consider two teacher architectures. \textbf{Privileged Teacher:} The privileged teacher $T_p$ considers only feature (privileged) information. Consequently, it is a simple ANN consisting of two layers. The first layer has 30 neurons employing the logistic activation function. The output of this layer is fed to the decision layer which is a simple 2-neuron dense SoftMax-activated layer. Once again a dropout of $0.1$ is applied before the last layer. 
\textbf{Fusion Teacher:} The fusion teacher $T_f$ has access to both privileged and prevalent information (i.e., features and frames). In particular, it fuses the penultimate layers of the frame and feature models described above. To account for the difference in dimensionality, an additional ReLu-activated layer of 30 neurons is added to the frame encoder to match the dimensions of the feature encoder. In this case the feature encoder is a single 30-neuron layer activated by ReLU. The last layer of the fusion encoder is a linear layer of 60 neurons activated by ReLU. The purpose of this last layer is to fuse information from both frames and features in order to yield the final representation. The output of the fusion layer is fed to a 2-neuron dense layer which is activated by SoftMax. A dropout of $0.1$ is applied before the output layer.  

\subsubsection{Baselines}\label{sec:lupi_ch_methodology_models_baseline}

There are two baseline architectures employed in this paper corresponding to student models that do not consider privileged information at all. The first architecture, used for comparison with the conventional LUPI framework, performs end-to-end affect classification from frames, bypassing the representation learning process. The second baseline, used for comparison with the PriCon framework, first trains the latent dimension of the student model via SCL and then probes the frozen learned embeddings with a SoftMax activated layer as the one used in all models of this work.   



\begin{figure}[!tb]
	\centering
    \subfloat[RECOLA]{\includegraphics[width=0.9\columnwidth]{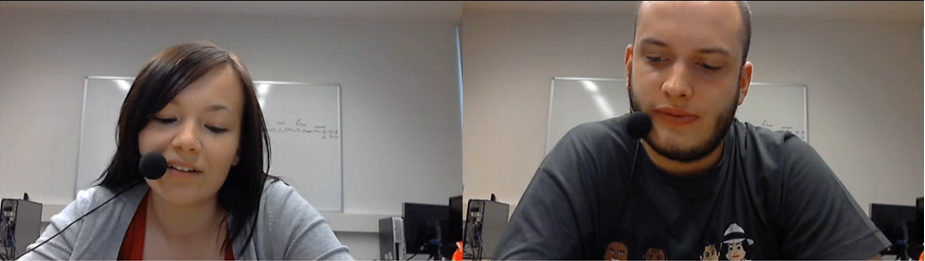}\label{fig:recola_samples}}\\
	\subfloat[Platformer Games (AGAIN)]{\includegraphics[width=0.9\columnwidth]{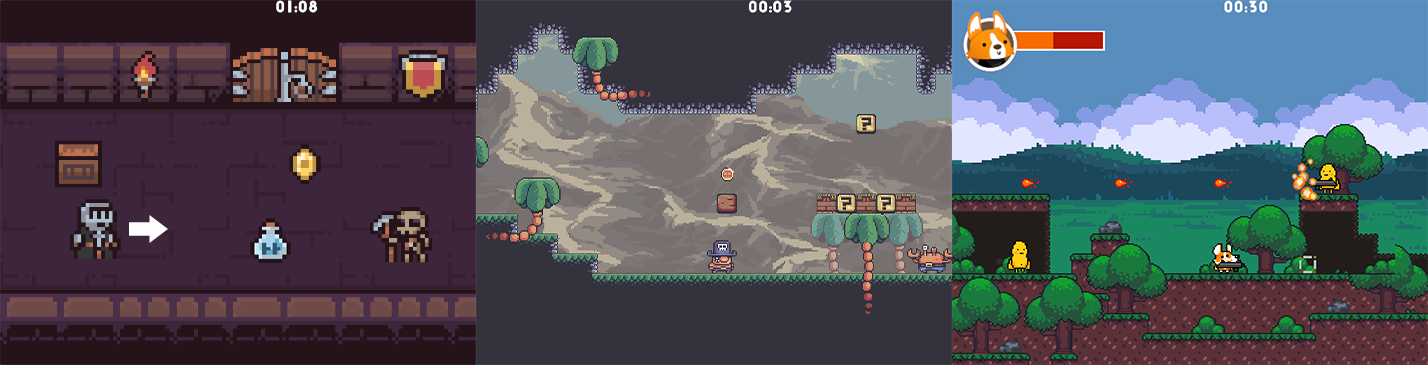}\label{fig:again_samples}}

	\vspace{-1em}
	\caption{ Sample frames from the datasets. (a) RECOLA: dyadic interactions. (b) AGAIN: three platformer games: \emph{Endless}, \emph{Pirates!}, and \emph{Run’N’Gun!} (from left to right).
		}\vspace{-1em}
	\label{fig:dataset_samples}
\end{figure}

\section{Datasets}

This section presents the datasets and preprocessing steps used to evaluate the performance of the proposed approach and test our hypotheses.

\subsection{RECOLA Database}

The RECOLA dataset (Figure \ref{fig:recola_samples}) contains 9.5 hours of synchronised multimodal data from 46 French-speaking participants, including audio, visual, and physiological signals (electrocardiogram and electrodermal activity). It features annotations for arousal and valence provided by six experts (third-person annotation). The dataset includes raw footage (frames) along with fine-grained feature sets: 40 visual features (e.g., facial action units, head pose, optical flow), 130 audio features (e.g., voice intensity, pitch, mel-frequency spectral coefficients), and 116 physiological features extracted from cardiac and electrodermal activity. Raw data is also available for further processing. Of the 46 participants, 34 consented to data sharing, 23 of which are included in the public release. 

\subsubsection{Processing RECOLA}\label{sec:scl_recola_preprocessing}

As mentioned earlier, the RECOLA database offers annotation traces for both arousal and valence. The same preprocessing approach is followed for both affective dimensions. Specifically, we segment each participant's session (features and frames) into overlapping time windows using a sliding step of \(400\,\mathrm{ms}\) and window lengths of \(1\), \(2\), and \(3\,\mathrm{s}\). These hyperparameters—the sliding step and window length—influence the size of the dataset and the temporal granularity of information contained in each window. Since the features and annotations are already synchronised, there is no need to account for the reaction time between the stimulus and the emotional response. After segmentation, each time window contains a sequence of feature vectors and a sequence of frames. To reduce computational complexity, we compute the average value for each feature within the window, representing the time window with a single feature vector. This ensures that the dimensionality of the feature vector is independent of the window length. For frame sequences, we retain \(5\) greyscale frames with dimensions \(224 \times 224\) per second. For example, the input for a \(3\,\mathrm{s}\) time window consists of a single feature vector (via averaging) and a frame tensor with dimensions \(15 \times 224 \times 224\). 

When it comes to affect annotation, we use the median annotation values per time window in order to derive a single trace segment---one per affect dimension---that mitigates inter-annotator disagreement \cite{grewe2007emotions}. The median affect trances are then averaged across time windows resulting in two scalar values per time window ($g_a$ for arousal and $g_v$ for valence). To produce the corresponding class labels we binarise the corresponding affect values. In RECOLA, the binarisation criterion is determined by the median ground truth value of the entire set of affect annotation traces (\(\tilde{g}_s\)) and a threshold \(\epsilon\). Specifically, a time window \(i\) is labelled as ``high" if \(g_{s_i} > \tilde{g}_s + \epsilon\) and as ``low" if \(g_{s_i} < \tilde{g}_s - \epsilon\). The median is used for RECOLA because each video stimulus has been annotated by the same six expert annotators, making the median a robust measure of central tendency. For RECOLA, \(\epsilon\) is set to 0.1, ensuring a precise exclusion of ambiguous annotations.

\subsection{AGAIN Dataset: Platformer Games}

For the experiments reported in this paper, we focus on the three games of the platformer genre featured in the AGAIN dataset (Figure \ref{fig:again_samples}) as they offer sufficiently diverse gameplay properties without in need of excessive computation for experimental validation. The three games examined include \textbf{Endless}, an infinite runner where players must avoid obstacles while automatically moving ever rightward, \textbf{Pirates!}, a jumping platformer similar to \emph{Super Mario Bros} (Nintendo, 1985), and \textbf{Run'N'Gun!}, a more complex game which requires players to move while aiming and shooting at enemies. All games have arcade-style controls of varying complexity, with Run'N'Gun! being the most complex of the three. Each game assigns a score to the player depending on their in-game performance. For the purpose of controlling the data collection process, we limit gameplay duration for all games to two minutes. The AGAIN-Platformers dataset consists of 120 participants that played and annotated their own gameplay in terms of arousal. Apart from raw gameplay footage the creators of AGAIN have also provided a high-level feature set for each game. The general feature set includes 14 features that describe player actions, environmental elements, and game events. Moreover, the dataset includes a set of game-specific extracted features for each game (33 for Endless!, 39 for Pirates! and 47 for Run’N’Gun!). The game-specific features correspond to player status (e.g., player health),
gameplay events (e.g., bot aims at player), bot status and the proximal and general game context (e.g., bot-player distance and pickups visible).

\subsubsection{Processing AGAIN}\label{sec:scl_again_preprocessing}

Each game in the AGAIN dataset provides self-reported arousal annotations, which are used in this work with a consistent preprocessing approach applied separately to each of the three games. Similarly to RECOLA, we segment each participant's session into overlapping time windows using a sliding step of \(500\,\mathrm{ms}\) and window lengths of \(1\), \(2\), and \(3\,\mathrm{s}\). Unlike the RECOLA dataset, synchronisation is required for AGAIN. To account for the reaction time between stimulus and emotional response, the arousal annotations are shifted backward by \(1\,\mathrm{s}\) \cite{pinitas2022rankneat,melhart2021towards}. Furthermore, the trace of each participant is normalised within $[0,1]$ due to the unbounded nature of the annotations in this dataset.

After preprocessing, each time window contains a sequence of feature vectors and a sequence of frames. To reduce computational load, we compute the average value of each feature within the window, resulting in a single feature vector that represents the entire window. For frame sequences, we retain \(5\) greyscale frames with dimensions \(224 \times 224\) per second. In the case of arousal annotation, there is no need to mitigate the disagreement between annotators since each participant provides a single trace per game per session (i.e., self-reporting annotations).  For each session, we compute the arousal score \(g_a\) as the mean value of this trace, yielding a scalar arousal score per session. This preprocessing pipeline is repeated independently for each of the three games in the dataset. For the AGAIN dataset, the binarisation criterion is based on the mean value of the annotation traces (\(\bar{g}_s\)) for each session, as it allows us to anchor binarisation relative to each participant's own reporting baseline, rather than applying a fixed global threshold that may not generalise across participants or game contexts \cite{pinitas2023predicting,makantasis2022invariant}. For AGAIN, \(\epsilon\) is set to $0.2$, accommodating the variability in self-reported annotations. These thresholds were empirically validated to ensure a balanced exclusion of ambiguous annotations while maintaining sufficient data \cite{makantasis2021pixels}. 

\section{Results}

This section presents the framework for evaluating the impact of contrastive teacher pretraining for privileged information on affect modelling and the experimental results obtained. 

\subsection{Evaluation Framework}
\label{ssec:evaluation_framework}

As mentioned earlier, we treat affect as a classification task. Hence, our models attempt to predict high and low affect (arousal or valence) states. We adopt a 5-fold cross-validation scheme to evaluate the models' performance across all experiments reported in the paper. When splitting a dataset, we make sure that data from the same participant is either in the training or the test set, not both. Before any train-test split we hold out $10\%$ of the participants randomly as our validation set. This set is used for stopping training early so as to avoid model overfitting; i.e., training stops after 5 epochs without loss improvement on the validation set. The training, validation and test sets are the same for all models reported. We report models' performance in terms of binary classification accuracy, since after class splitting, the datasets are balanced. A key component of the LUPI training objective is the hyperparameter $\alpha$, which controls the teacher's influence. We perform an exhaustive search over candidate values $\alpha \in \{0.25, 0.5, 0.75, 1.0\}$ using the validation set and adopt the value that yields the best performance for subsequent experiments. All reported results are statistically validated via pairwise t-tests ($p<0.05$).

In this work, student models make predictions using solely information that is available in the wild; in our experiments that is the raw footage frames. The student models trained via LUPI are denoted as $S_p$ and $S_f$ and have access to privileged information through the Privileged Teacher and the Fusion Teacher, respectively. We compare the student's performance against the performance achieved by the Fusion Teacher ($T_f$) that uses all modalities (i.e., features \& frames) for training and testing, and the Privileged Teacher ($T_p$) that makes predictions using only privileged information. $E$ refers to models that do not utilise privileged information during training. In Tables \ref{tab:res_lupi_recola} and \ref{tab:res_lupi_again} the LUPI columns refer to the conventional approach where the teachers have been trained in an end-to-end fashion in this case $E$ is also trained in the same manner. The PC column refers to LUPI with teachers that have been pretrained via SCL (proposed PriCon framework). The accuracy of the $T_p$, $T_f$ and $E$ models in the PC setting is obtained by applying a linear probe on top of the learned representations. 

\subsection{The Importance of Privileged Information}\label{sec:lupi_ch_results_best}



\begin{table}[!tb]
	\centering
	\caption{RECOLA: Average accuracy prediction for arousal and valence. $E$ corresponds to baseline models. $S_p$ and $S_f$ correspond to student models that have been trained under the privileged ($T_p$) and fusion ($T_f$) teacher, respectively. LUPI and PC denote, respectively, vanilla learning via privileged information and our introduced PriCon framework. Bold values correspond to the student model with the highest accuracy and underlined values correspond to models whose performance is statistically on par ($p\ge 0.05$) with that model.}
   
	\resizebox{\columnwidth}{!}{
    \begin{tabular}{c||c|c||c|c||c|c}
		\textbf{Arousal} & 
		\multicolumn{2}{c}{\textbf{1 second}} &
		\multicolumn{2}{c}{\textbf{2 seconds}} &
		\multicolumn{2}{c}{\textbf{3 seconds}} \\ 
		\hline \hline
		Model & LUPI & PC & LUPI & PC & LUPI & PC \\
        \hline
        $E$ & 59.41 & 60.60 & 55.73 & 57.86 & 58.22 & 60.85\\
        \hline
		$S_p$ & 63.61 & \textbf{65.27} & 60.68 & 60.64 & 60.96 & \textbf{63.07}\\
            $S_f$ & 60.75 & 63.74 & 61.20 & \textbf{62.31} & 61.62 & \underline{62.31}\\
            \hline
            $T_p$ & 69.24 & 71.41 & 71.59 & 73.58 & 73.11 & 75.45\\
            $T_f$ & 63.04 & \underline{64.60} & 66.73 & 68.86 & 66.22 & 69.49\\
		\hline \hline 
	
    \textbf{Valence} & 
		\multicolumn{2}{c}{\textbf{1 second}} &
		\multicolumn{2}{c}{\textbf{2 seconds}} &
		\multicolumn{2}{c}{\textbf{3 seconds}} \\ 
		\hline \hline
		Model & LUPI & PC & LUPI & PC & LUPI & PC \\
        \hline
        $E$ & \underline{60.84} & \underline{60.19} & \underline{60.60} & \underline{60.40} & 62.36 & 61.34\\
        \hline
		$S_p$ & \underline{59.94} & \textbf{62.26} & \underline{60.07} & \underline{59.82} & \textbf{64.90} & 62.42\\
            $S_f$ & \underline{61.66} & \underline{61.58} & \textbf{60.78} & \underline{60.09} & 62.51 & 59.61\\
            \hline
            $T_p$ & \underline{60.95} & \underline{62.10} & 56.51 & \underline{59.48} & \underline{65.15} & \underline{62.80}\\
            $T_f$ & \underline{62.74} & \underline{62.38} & \underline{61.26} & \underline{61.99} & \underline{64.74} & \underline{64.21}\\

    \end{tabular}
    }
	\label{tab:res_lupi_recola}

\end{table}

\begin{table}[!tb]
	\centering
	\caption{AGAIN: Average accuracy prediction for arousal across three platformer games of the dataset. 
    Bold values correspond to the student model with the highest accuracy and underlined values correspond to models whose performance is statistically on par ($p\ge 0.05$) with that model.} 
    \resizebox{\columnwidth}{!}{
	\begin{tabular}{c||c|c||c|c||c|c}
		\textbf{Endless} & 
		\multicolumn{2}{c}{\textbf{1 second}} &
		\multicolumn{2}{c}{\textbf{2 seconds}} &
		\multicolumn{2}{c}{\textbf{3 seconds}} \\ 
		\hline \hline
		Model & LUPI & PC & LUPI & PC & LUPI & PC \\
        \hline
        $E$ & 65.96 & \underline{67.83} & 63.14 & 65.84 & 65.52 & 65.54\\
        \hline
		$S_p$ & 66.71 & \textbf{67.95} & 67.08 & \textbf{68.21} & 66.36 & \textbf{69.51}\\
            $S_f$ & 66.17 & \underline{67.26} & 66.07 & \underline{67.50} & 66.27 & \underline{69.29}\\
            \hline
            $T_p$ & 71.57 & 72.65 & 73.25 & 74.34 & 72.77 & 81.33\\
            $T_f$ & 71.16 & 72.83 & 73.14 & 74.66 & 72.06 & 78.95\\         
		\hline \hline 
	
    \textbf{Pirates!} & 
		\multicolumn{2}{c}{\textbf{1 second}} &
		\multicolumn{2}{c}{\textbf{2 seconds}} &
		\multicolumn{2}{c}{\textbf{3 seconds}} \\ 
		\hline \hline
		Model & LUPI & PC & LUPI & PC & LUPI & PC \\
        \hline
        
        $E$ & \underline{67.63} & \underline{67.03} & 66.22 & 65.49 & \underline{66.54} & \underline{66.57}\\
        \hline
		$S_p$ & \underline{66.79} & \underline{66.89} & 66.03 & \underline{67.06} & 66.42 & \underline{66.86}\\
            $S_f$ & \underline{66.60} &\textbf{ 67.12} & \underline{66.89} & \textbf{67.25} & 66.01 & \textbf{67.39}\\
            \hline
            $T_p$ & 65.36 & \underline{67.46} & 65.50 & \underline{67.05} & 65.66 & \underline{66.72}\\
            $T_f$ & 65.66 & \underline{67.16} & 65.62 & \underline{67.75} & 65.65 & 66.46\\       
    
    \hline \hline 
	
    \textbf{Run'N'Gun!} & 
		\multicolumn{2}{c}{\textbf{1 second}} &
		\multicolumn{2}{c}{\textbf{2 seconds}} &
		\multicolumn{2}{c}{\textbf{3 seconds}} \\ 
		\hline \hline
		Model & LUPI & PC & LUPI & PC & LUPI & PC \\        \hline
        $E$ & 67.40 & 70.21 & 68.11 & 69.95 & 68.07 & 68.77\\
        \hline
		$S_p$ & 70.40 & \textbf{71.32} & 70.43 & 71.41 & 69.73 & \textbf{71.11}\\
            $S_f$ & \underline{70.66} & \underline{71.21} & 69.80 & \textbf{72.05} & 68.96 & \underline{70.43}\\
            \hline
            $T_p$ & 73.05 & 74.66 & 74.15 & 75.14 & 73.25 & 73.68\\
            $T_f$ & \underline{71.40} & 73.43 & \underline{72.11} & 73.95 & 73.07 & 73.88\\
    
    \end{tabular}
    }
	\label{tab:res_lupi_again}
\end{table}

To test \textbf{H1}---which posits that LUPI can enhance affect detection under real-world (in-vivo) conditions, producing performances comparable to in-vitro models---we evaluate and compare the performance of baseline, teacher, and student models within the LUPI framework. Table~\ref{tab:res_lupi_recola} reports the average 5-fold validation accuracy for high–low arousal and valence classification tasks on the RECOLA dataset. The results are analysed across three different time windows (1s, 2s, 3s). The models evaluated include baseline models, teacher models (Fusion and Privileged Teachers), and student models trained via the LUPI framework (see LUPI column).

Baseline models ($E$) consistently achieve low accuracy scores across all six experimental settings, indicating that models relying solely on raw frame data are unable to achieve competitive performance. This highlights the limitations of training models without incorporating privileged information or additional learning signals. The teacher models—namely the Fusion Teacher ($T_f$) and the Privileged Teacher ($T_p$)—yield significantly higher accuracy values for arousal compared to the baseline models. $T_p$ teachers, which consider only privileged features, consistently achieve the highest accuracy, establishing the upper bound for arousal classification. On the other hand, $T_f$ teachers—trained on both privileged information and frames—achieve competitive results but typically perform worse than $T_p$.

LUPI-trained student models ($S_p$, $S_f$) demonstrate substantial improvements over baseline models. Specifically, $S_f$, which is trained under the supervision of the Fusion Teacher, performs on par with or even outperforms $S_p$, which is guided by the Privileged Teacher. This finding suggests that the inclusion of frames in the $T_f$ teacher can allow for better information transfer during the LUPI training process.

While significant improvements are observed for arousal classification, valence classification results remain relatively stable across models, training strategies, and time window lengths. This can be attributed to the inherent challenges of valence prediction, which relies on subtle and ambiguous cues that are less dynamic and harder to capture than arousal. Valence-related expressions often appear visually similar across emotional contexts, making it difficult for models using raw frames to learn discriminative features~\cite{aslam2023privileged,makantasis2023learning}. Additionally, while privileged information provides strong supervisory signals for arousal, it may not capture sufficient valence-specific information, leading to weaker guidance from the $T_p$ and $T_f$ teachers. This suggests that valence recognition might require additional contextual information, such as semantics or interaction dynamics~\cite{souter2023valence,wieser2012faces}.

Table~\ref{tab:res_lupi_again} presents the arousal classification results for the AGAIN dataset, which includes three platformer games: \textit{Run'N'Gun!}, \textit{Pirates!}, and \textit{Endless}. Each game is evaluated for time windows of 1s, 2s, and 3s. In \textit{Run'N'Gun!}, $T_p$ achieves the highest performance, establishing an upper bound. Among the LUPI-trained student models, $S_p$ performs on par with or slightly better $S_f$, and both improve upon the baseline models ($E$).

For \textit{Pirates!}, performance remains relatively stable across all time windows, with limited improvements as the temporal context increases. This stability suggests that arousal cues are less dynamic and harder to capture in this game due to subtler changes in play. In this case, both teachers perform on par; the same holds for $S_f$ and $S_p$ among the student models. It is worth noting that here, teacher models perform on par with student models that do not use privileged information. Consequently, there is no information gain to be exploited by the LUPI framework. Finally, in \textit{Endless}, the end-to-end teachers perform on par. Among the LUPI-trained students, $S_f$ and $S_p$ also perform similarly across all time windows, consistently improving over the baseline models $E$.

The obtained results collectively across two different datasets and 3 different affective dimensions validate \textbf{H1}, confirming that the LUPI framework can enhance affect detection under real-world (in-vivo) conditions. LUPI-trained student models consistently outperform baseline models and, in several cases, closely approximate the performance of their privileged teacher counterparts. This demonstrates the potential of privileged supervision to guide the learning of compact, frame-only student models that generalise well in realistic settings where privileged modalities are unavailable at test time. 

\subsection{Privileged Contrastive Pretraining}

To evaluate \textbf{H2}, we investigate the impact of PriCon---a framework that integrates SCL within LUPI---on the performance of both teacher and student models. The central question is whether teacher pretraining via SCL leads to more effective student learning and ultimately enhances binary affect classification. To evaluate this, we focus on the results reported in the `PC' columns of Tables~\ref{tab:res_lupi_recola} and~\ref{tab:res_lupi_again}, which isolate the influence of PriCon on model performance across different datasets, affective states, and time window configurations.

For the RECOLA dataset (Table~\ref{tab:res_lupi_recola}), PriCon significantly improves model performance across the board. SCL-pretrained teacher models, ${T_f}$ and ${T_p}$, (PC column) not only match but often outperform their end-to-end LUPI counterparts. This indicates that contrastive pretraining enables the learning of more discriminative and generalisable affect representations, which are beneficial regardless of whether the teacher is trained on fused or privileged data. Importantly, these benefits are transferred downstream: student models ${S_p}$ and ${S_f}$ consistently outperform the end-to-end baseline $E$, and in multiple cases even match or exceed the performance of SCL-pretrained $E$. These results validate the effectiveness of the PriCon framework, allowing student models to benefit from the increased performace of the SCL pretrained teachers. Valence classification results, however, remain relatively stable. While PriCon provides modest gains, the improvements are less substantial due to the subtle and ambiguous nature of valence cues in visual data, as discussed in Section~\ref{sec:lupi_ch_results_best}.

In the AGAIN dataset (Table~\ref{tab:res_lupi_again}), PriCon continues to show strong and consistent gains for student models. In \textit{Run'N'Gun!}, both ${S_p}$ and ${S_f}$ outperform not only the baseline $E$ but also their end-to-end LUPI versions, reinforcing the utility of combining contrastive objectives (for teacher pretraining) with the conventional LUPI paradigm. This demonstrates PriCon's capacity to bridge the modality gap during student training and produce better generalising models. In \textit{Pirates!}, although the overall gains are smaller, PriCon yields noticeable improvements in 2-second time windows. Here, ${S_p}$ and ${S_f}$ significantly outperform $E$, even when teacher performance remains similar. In \textit{Endless}, PriCon teachers train students (see PC column) that achieve higher accuracy than both their end-to-end counterparts (LUPI column) and the baseline models $E$. Notably, the performance gap is most pronounced for the 2s and 3s windows, suggesting that temporal context plays a role in enhancing the benefits of SCL teacher pretraining. 

In summary, obtained results validate \textbf{H2}, demonstrating the effectiveness of the PriCon framework. By integrating SCL for teacher pretraining and LUPI for privileged information distillation, PriCon enables the development of robust affective models with improve generalisation across tasks (affect dimensions), input modalities, and temporal resolutions. 



	

\section{Discussion}

The findings presented in this paper underscore the significant potential of \emph{privileged contrastive pretraining} for affect modelling. By focusing on binary classification (high vs. low affective state) our introduced method demonstrated improved robustness and performance across both RECOLA and AGAIN datasets. However, several methodological limitations and considerations must be acknowledged. Although arousal classification benefits substantially from PriCon, valence results show only marginal gains, indicating that valence remains a harder target for generalisation. Thus future work should explore the integration of user context \cite{tammewar2019modeling} and MAMBA \cite{liang2025mamba} architectures within the PriCon framework to further improve valence prediction.

The study employed an exhaustive search protocol to fine tune the $\alpha$ hyperparameter, which controls the influence of the teacher in LUPI. While this approach ensured optimal performance for the chosen datasets, it may not generalise well to other datasets or applications. The fixed hyperparameter tuning approach assumes a universal solution, potentially missing dataset-specific variations. Tuning was done for 1-second time windows, assuming the optimal $\alpha$ value is consistent across all windows. While more efficient methods like Bayesian optimisation could improve tuning, their computational demands were deemed impractical for this study. Future work could explore adaptive optimisation techniques such as gradient-based hyperparameter learning \cite{maclaurin2015gradient,franceschi2017forward} for better efficiency and performance across datasets.

Beyond hyperparameter tuning, the manual selection of privileged features represents another limitation of this study. While the selected features were thoughtfully aligned with domain knowledge and task-specific objectives, the ad-hoc approach followed may have overlooked complex, latent structures within the datasets that automated methods could exploit. Attention mechanisms and feature-ranking algorithms \cite{gui2019afs,yoon2005feature} could help identify and prioritise the most relevant features, revealing patterns missed by manual selection. However, implementing these methods would require additional resources, potentially diverting attention from the study's main goal of evaluating the proposed framework. While manual selection ensured interpretability, future work could explore automated techniques to enhance scalability and effectiveness.

While the binary classification paradigm is widely used to showcase the robustness of AC frameworks \cite{sanchez2022machine,pinitas2024varying,khare2024emotion,kumar2023classification} , a natural extension would be to explore the potential of PriCon in semi-supervised and self-supervised learning paradigms. Such extensions could unlock the ability to learn powerful general-purpose representations that are better suited for diverse affect modelling tasks, especially in scenarios with limited labelled data. These paradigms would be particularly beneficial for scaling affective computing systems in real-world applications where data labelling is often costly and time-consuming.

Another promising direction involves extending PriCon to ranking paradigms, which are essential for ordinal affect modelling tasks \citep{makantasis2021affranknet+,yannakakis2018ordinal}. The incorporation of PriCon into these paradigms could enable models to better exploit privileged information  for inferring subtle affective gradients, thereby improving their ability to model subjective human experiences. Furthermore, combining PriCon with few-shot learning strategies \cite{wang2020generalizing, pinitas2024silhouette} could facilitate robust affect modelling in low-data regimes, a common challenge in real-world scenarios. However, these extensions would require significant methodological adaptations. These adaptations---while outside the scope of the current study---represent exciting opportunities for advancing the field and addressing more complex affective computing challenges in future research. Finally, regarding evaluation, this study reports binary classification accuracy since the datasets are balanced. Nonetheless, future work could explore alternative metrics---such as F1 score, AUC, or mean absolute error---that may offer complementary insights, particularly in the presence of class imbalance or in regression-based affect modelling.

\section{Conclusions}

In this paper, we introduced a methodology for affect modelling in real-world scenarios by leveraging privileged information and teacher pre-training via SCL. Our central hypothesis posits that privileged information can facilitate the reliable transfer of affect models from controlled environments, where abundant high-quality data is available, to real-world settings. We evaluated this hypothesis in the RECOLA and AGAIN datasets, which include both raw visual data and high-level handcrafted features. We considered all handcrafted features as privileged information that is accessible solely during model training, while also considering raw visual data such as video frames as available during both training and testing. As part of the PriCon framework, we pretrain teacher models via SCL and subsequently transfer their knowledge to student models, which operate without privileged information during testing. This approach assumes that teachers with a higher predictive power can produce more robust student models.

Our findings for arousal and valence prediction highlight the combined advantages of leveraging privileged information and contrastive pretraining. Affect models trained via the PriCon framework not only match but often exceed the performance of their teacher models, while operating solely on frame-based features at test time. Remarkably, in many cases, PriCon models achieve performance comparable to models trained with access to all modalities during both training and testing. The proposed methodology has broad applicability to affective computing tasks involving multimodal data. It is especially valuable in scenarios where access to certain modalities is limited or unavailable, offering a scalable and effective approach to emotion recognition in the wild. The findings underscore the potential of PriCon as a paradigm towards further bridging the gap between \emph{in-vitro} and \emph{in-vivo} affective modelling, offering a scalable and practical solution for real-world applications.

\section*{Safe and Responsible Innovation Statement}
The experiments presented in this paper were conducted using publicly available datasets containing frames, fine-grained features, and affect annotations. Any personally identifiable information was anonymised with untraceable IDs, ensuring that such data was not accessible to us. Participants consented to having their faces publicly available. Furthermore, the datasets used do not contain potentially offensive content, and the proposed framework is based on open-source methods, ensuring reproducibility. Finally, to the best of our knowledge, our work does not contribute to the development of deceptive applications or exacerbate existing privacy or discriminatory concerns.

\section*{Acknowledgements}
This is the author's version of the work. It is posted here for your personal use. Not for redistribution. The definitive version will be published in Proceedings of the  27th International Conference on Multimodal Interaction (ICMI '25), 
\url{https://doi.org/10.1145/3716553.3750766}.
This project has received funding from the Malta Council for Science
and Technology through the SINO-MALTA Fund 2022, Project OPtiMaL.

\bibliographystyle{ACM-Reference-Format}
\bibliography{sample-base}


\begin{thebibliography}{46}


\ifx \showCODEN    \undefined \def \showCODEN     #1{\unskip}     \fi
\ifx \showISBNx    \undefined \def \showISBNx     #1{\unskip}     \fi
\ifx \showISBNxiii \undefined \def \showISBNxiii  #1{\unskip}     \fi
\ifx \showISSN     \undefined \def \showISSN      #1{\unskip}     \fi
\ifx \showLCCN     \undefined \def \showLCCN      #1{\unskip}     \fi
\ifx \shownote     \undefined \def \shownote      #1{#1}          \fi
\ifx \showarticletitle \undefined \def \showarticletitle #1{#1}   \fi
\ifx \showURL      \undefined \def \showURL       {\relax}        \fi
\providecommand\bibfield[2]{#2}
\providecommand\bibinfo[2]{#2}
\providecommand\natexlab[1]{#1}
\providecommand\showeprint[2][]{arXiv:#2}

\bibitem[Aslam et~al\mbox{.}(2023)]%
        {aslam2023privileged}
\bibfield{author}{\bibinfo{person}{Muhammad~Haseeb Aslam}, \bibinfo{person}{Muhammad~Osama Zeeshan}, \bibinfo{person}{Marco Pedersoli}, \bibinfo{person}{Alessandro~L Koerich}, \bibinfo{person}{Simon Bacon}, {and} \bibinfo{person}{Eric Granger}.} \bibinfo{year}{2023}\natexlab{}.
\newblock \showarticletitle{Privileged knowledge distillation for dimensional emotion recognition in the wild}. In \bibinfo{booktitle}{\emph{Proceedings of the IEEE/CVF conference on computer vision and pattern recognition}}. \bibinfo{pages}{3338--3347}.
\newblock


\bibitem[Botelho et~al\mbox{.}(2017)]%
        {botelho_improving_2017}
\bibfield{author}{\bibinfo{person}{Anthony~F. Botelho}, \bibinfo{person}{Ryan~S. Baker}, {and} \bibinfo{person}{Neil~T. Heffernan}.} \bibinfo{year}{2017}\natexlab{}.
\newblock \showarticletitle{Improving {Sensor}-{Free} {Affect} {Detection} {Using} {Deep} {Learning}}. In \bibinfo{booktitle}{\emph{Artificial {Intelligence} in {Education}}} \emph{(\bibinfo{series}{Lecture {Notes} in {Computer} {Science}})}, \bibfield{editor}{\bibinfo{person}{Elisabeth André}, \bibinfo{person}{Ryan Baker}, \bibinfo{person}{Xiangen Hu}, \bibinfo{person}{Ma. Mercedes~T. Rodrigo}, {and} \bibinfo{person}{Benedict du~Boulay}} (Eds.). \bibinfo{publisher}{Springer International Publishing}, \bibinfo{address}{Cham}, \bibinfo{pages}{40--51}.
\newblock
\showISBNx{978-3-319-61425-0}
\href{https://doi.org/10.1007/978-3-319-61425-0_4}{doi:\nolinkurl{10.1007/978-3-319-61425-0_4}}


\bibitem[Franceschi et~al\mbox{.}(2017)]%
        {franceschi2017forward}
\bibfield{author}{\bibinfo{person}{Luca Franceschi}, \bibinfo{person}{Michele Donini}, \bibinfo{person}{Paolo Frasconi}, {and} \bibinfo{person}{Massimiliano Pontil}.} \bibinfo{year}{2017}\natexlab{}.
\newblock \showarticletitle{Forward and reverse gradient-based hyperparameter optimization}. In \bibinfo{booktitle}{\emph{International conference on machine learning}}. PMLR, \bibinfo{pages}{1165--1173}.
\newblock


\bibitem[Grewe et~al\mbox{.}(2007)]%
        {grewe2007emotions}
\bibfield{author}{\bibinfo{person}{Oliver Grewe}, \bibinfo{person}{Frederik Nagel}, \bibinfo{person}{Reinhard Kopiez}, {and} \bibinfo{person}{Eckart Altenm{\"u}ller}.} \bibinfo{year}{2007}\natexlab{}.
\newblock \showarticletitle{Emotions over time: synchronicity and development of subjective, physiological, and facial affective reactions to music.}
\newblock \bibinfo{journal}{\emph{Emotion}} \bibinfo{volume}{7}, \bibinfo{number}{4} (\bibinfo{year}{2007}), \bibinfo{pages}{774}.
\newblock


\bibitem[Gui et~al\mbox{.}(2019)]%
        {gui2019afs}
\bibfield{author}{\bibinfo{person}{Ning Gui}, \bibinfo{person}{Danni Ge}, {and} \bibinfo{person}{Ziyin Hu}.} \bibinfo{year}{2019}\natexlab{}.
\newblock \showarticletitle{AFS: An attention-based mechanism for supervised feature selection}. In \bibinfo{booktitle}{\emph{Proceedings of the AAAI conference on artificial intelligence}}, Vol.~\bibinfo{volume}{33}. \bibinfo{pages}{3705--3713}.
\newblock


\bibitem[Hinton et~al\mbox{.}(2015)]%
        {hinton2015distilling}
\bibfield{author}{\bibinfo{person}{Geoffrey Hinton}, \bibinfo{person}{Oriol Vinyals}, {and} \bibinfo{person}{Jeff Dean}.} \bibinfo{year}{2015}\natexlab{}.
\newblock \showarticletitle{Distilling the Knowledge in a Neural Network}.
\newblock \bibinfo{journal}{\emph{Stat}}  \bibinfo{volume}{1050} (\bibinfo{year}{2015}).
\newblock


\bibitem[Khare et~al\mbox{.}(2024)]%
        {khare2024emotion}
\bibfield{author}{\bibinfo{person}{Smith~K Khare}, \bibinfo{person}{Victoria Blanes-Vidal}, \bibinfo{person}{Esmaeil~S Nadimi}, {and} \bibinfo{person}{U~Rajendra Acharya}.} \bibinfo{year}{2024}\natexlab{}.
\newblock \showarticletitle{Emotion recognition and artificial intelligence: A systematic review (2014--2023) and research recommendations}.
\newblock \bibinfo{journal}{\emph{Information fusion}}  \bibinfo{volume}{102} (\bibinfo{year}{2024}), \bibinfo{pages}{102019}.
\newblock


\bibitem[Kumar et~al\mbox{.}(2023)]%
        {kumar2023classification}
\bibfield{author}{\bibinfo{person}{Gs~Shashi Kumar}, \bibinfo{person}{Niranjana Sampathila}, {and} \bibinfo{person}{Roshan~Joy Martis}.} \bibinfo{year}{2023}\natexlab{}.
\newblock \showarticletitle{Classification of human emotional states based on valence-arousal scale using electroencephalogram}.
\newblock \bibinfo{journal}{\emph{Journal of Medical Signals \& Sensors}} \bibinfo{volume}{13}, \bibinfo{number}{2} (\bibinfo{year}{2023}), \bibinfo{pages}{173--182}.
\newblock


\bibitem[Li et~al\mbox{.}(2020)]%
        {li_how_2020}
\bibfield{author}{\bibinfo{person}{Lin Li}, \bibinfo{person}{Tiong-Thye Goh}, {and} \bibinfo{person}{Dawei Jin}.} \bibinfo{year}{2020}\natexlab{}.
\newblock \showarticletitle{How textual quality of online reviews affect classification performance: a case of deep learning sentiment analysis}.
\newblock \bibinfo{journal}{\emph{Neural Computing and Applications}} \bibinfo{volume}{32}, \bibinfo{number}{9} (\bibinfo{date}{May} \bibinfo{year}{2020}), \bibinfo{pages}{4387--4415}.
\newblock
\href{https://doi.org/10.1007/s00521-018-3865-7}{doi:\nolinkurl{10.1007/s00521-018-3865-7}}


\bibitem[Li et~al\mbox{.}(2021)]%
        {li2021contrastive}
\bibfield{author}{\bibinfo{person}{Mao Li}, \bibinfo{person}{Bo Yang}, \bibinfo{person}{Joshua Levy}, \bibinfo{person}{Andreas Stolcke}, \bibinfo{person}{Viktor Rozgic}, \bibinfo{person}{Spyros Matsoukas}, \bibinfo{person}{Constantinos Papayiannis}, \bibinfo{person}{Daniel Bone}, {and} \bibinfo{person}{Chao Wang}.} \bibinfo{year}{2021}\natexlab{}.
\newblock \showarticletitle{Contrastive unsupervised learning for speech emotion recognition}. In \bibinfo{booktitle}{\emph{ICASSP 2021-2021 IEEE International Conference on Acoustics, Speech and Signal Processing (ICASSP)}}. IEEE, \bibinfo{pages}{6329--6333}.
\newblock


\bibitem[Liang et~al\mbox{.}(2025)]%
        {liang2025mamba}
\bibfield{author}{\bibinfo{person}{Yuheng Liang}, \bibinfo{person}{Zheyu Wang}, \bibinfo{person}{Feng Liu}, \bibinfo{person}{Mingzhou Liu}, {and} \bibinfo{person}{Yu Yao}.} \bibinfo{year}{2025}\natexlab{}.
\newblock \showarticletitle{Mamba-va: A mamba-based approach for continuous emotion recognition in valence-arousal space}. In \bibinfo{booktitle}{\emph{Proceedings of the Computer Vision and Pattern Recognition Conference}}. \bibinfo{pages}{5651--5656}.
\newblock


\bibitem[Liu et~al\mbox{.}(2022)]%
        {liu2022social}
\bibfield{author}{\bibinfo{person}{Huan Liu}, \bibinfo{person}{Ke Li}, \bibinfo{person}{Jianping Fan}, \bibinfo{person}{Caixia Yan}, \bibinfo{person}{Tao Qin}, {and} \bibinfo{person}{Qinghua Zheng}.} \bibinfo{year}{2022}\natexlab{}.
\newblock \showarticletitle{Social Image-text Sentiment Classification With Cross-Modal Consistency and Knowledge Distillation}.
\newblock \bibinfo{journal}{\emph{IEEE Transactions on Affective Computing}} (\bibinfo{year}{2022}).
\newblock


\bibitem[Lopez-Paz et~al\mbox{.}(2016)]%
        {lopez2016unifying}
\bibfield{author}{\bibinfo{person}{David Lopez-Paz}, \bibinfo{person}{L{\'e}on Bottou}, \bibinfo{person}{Bernhard Sch{\"o}lkopf}, {and} \bibinfo{person}{Vladimir Vapnik}.} \bibinfo{year}{2016}\natexlab{}.
\newblock \showarticletitle{Unifying distillation and privileged information}.
\newblock  (\bibinfo{year}{2016}).
\newblock


\bibitem[Maclaurin et~al\mbox{.}(2015)]%
        {maclaurin2015gradient}
\bibfield{author}{\bibinfo{person}{Dougal Maclaurin}, \bibinfo{person}{David Duvenaud}, {and} \bibinfo{person}{Ryan Adams}.} \bibinfo{year}{2015}\natexlab{}.
\newblock \showarticletitle{Gradient-based hyperparameter optimization through reversible learning}. In \bibinfo{booktitle}{\emph{International conference on machine learning}}. PMLR, \bibinfo{pages}{2113--2122}.
\newblock


\bibitem[Mai et~al\mbox{.}(2022)]%
        {mai2022hybrid}
\bibfield{author}{\bibinfo{person}{Sijie Mai}, \bibinfo{person}{Ying Zeng}, \bibinfo{person}{Shuangjia Zheng}, {and} \bibinfo{person}{Haifeng Hu}.} \bibinfo{year}{2022}\natexlab{}.
\newblock \showarticletitle{Hybrid contrastive learning of tri-modal representation for multimodal sentiment analysis}.
\newblock \bibinfo{journal}{\emph{IEEE Transactions on Affective Computing}} \bibinfo{volume}{14}, \bibinfo{number}{3} (\bibinfo{year}{2022}), \bibinfo{pages}{2276--2289}.
\newblock


\bibitem[Makantasis(2021)]%
        {makantasis2021affranknet+}
\bibfield{author}{\bibinfo{person}{Konstantinos Makantasis}.} \bibinfo{year}{2021}\natexlab{}.
\newblock \showarticletitle{Affranknet+: ranking affect using privileged information}. In \bibinfo{booktitle}{\emph{2021 9th International Conference on Affective Computing and Intelligent Interaction Workshops and Demos (ACIIW)}}. IEEE, \bibinfo{pages}{1--8}.
\newblock


\bibitem[Makantasis et~al\mbox{.}(2021a)]%
        {makantasis2021pixels}
\bibfield{author}{\bibinfo{person}{Konstantinos Makantasis}, \bibinfo{person}{Antonios Liapis}, {and} \bibinfo{person}{Georgios~N Yannakakis}.} \bibinfo{year}{2021}\natexlab{a}.
\newblock \showarticletitle{The Pixels and Sounds of Emotion: General-Purpose Representations of Arousal in Games}.
\newblock \bibinfo{journal}{\emph{IEEE Trans. on Affective Computing}} (\bibinfo{year}{2021}).
\newblock


\bibitem[Makantasis et~al\mbox{.}(2021b)]%
        {makantasis2021privileged}
\bibfield{author}{\bibinfo{person}{Konstantinos Makantasis}, \bibinfo{person}{David Melhart}, \bibinfo{person}{Antonios Liapis}, {and} \bibinfo{person}{Georgios~N Yannakakis}.} \bibinfo{year}{2021}\natexlab{b}.
\newblock \showarticletitle{Privileged Information for Modeling Affect In The Wild}. In \bibinfo{booktitle}{\emph{Proc. of the IEEE Int. Conf. on Affective Computing and Intelligent Interaction}}.
\newblock


\bibitem[Makantasis et~al\mbox{.}(2022)]%
        {makantasis2022invariant}
\bibfield{author}{\bibinfo{person}{Konstantinos Makantasis}, \bibinfo{person}{Kosmas Pinitas}, \bibinfo{person}{Antonios Liapis}, {and} \bibinfo{person}{Georgios~N Yannakakis}.} \bibinfo{year}{2022}\natexlab{}.
\newblock \showarticletitle{The Invariant Ground Truth of Affect}. In \bibinfo{booktitle}{\emph{2022 10th International Conference on Affective Computing and Intelligent Interaction Workshops and Demos (ACIIW)}}. IEEE, \bibinfo{pages}{1--8}.
\newblock


\bibitem[Makantasis et~al\mbox{.}(2023)]%
        {makantasis2023learning}
\bibfield{author}{\bibinfo{person}{Konstantinos Makantasis}, \bibinfo{person}{Kosmas Pinitas}, \bibinfo{person}{Antonios Liapis}, {and} \bibinfo{person}{Georgios~N Yannakakis}.} \bibinfo{year}{2023}\natexlab{}.
\newblock \showarticletitle{From the lab to the wild: Affect modeling via privileged information}.
\newblock \bibinfo{journal}{\emph{IEEE Transactions on Affective Computing}} (\bibinfo{year}{2023}).
\newblock


\bibitem[Melhart et~al\mbox{.}(2021a)]%
        {melhart2021again}
\bibfield{author}{\bibinfo{person}{David Melhart}, \bibinfo{person}{Antonios Liapis}, {and} \bibinfo{person}{Georgios~N. Yannakakis}.} \bibinfo{year}{2021}\natexlab{a}.
\newblock \showarticletitle{{The Affect Game AnnotatIoN} ({AGAIN}) Dataset}.
\newblock \bibinfo{journal}{\emph{arXiv preprint arXiv:2104.02643}} (\bibinfo{year}{2021}).
\newblock
\showeprint[arxiv]{2104.02643}~[cs.HC]


\bibitem[Melhart et~al\mbox{.}(2021b)]%
        {melhart2021towards}
\bibfield{author}{\bibinfo{person}{David Melhart}, \bibinfo{person}{Antonios Liapis}, {and} \bibinfo{person}{Georgios~N Yannakakis}.} \bibinfo{year}{2021}\natexlab{b}.
\newblock \showarticletitle{Towards general models of player experience: A study within genres}. In \bibinfo{booktitle}{\emph{2021 IEEE Conference on Games (CoG)}}. IEEE, \bibinfo{pages}{01--08}.
\newblock


\bibitem[Pinitas et~al\mbox{.}(2022a)]%
        {pinitas2022rankneat}
\bibfield{author}{\bibinfo{person}{Kosmas Pinitas}, \bibinfo{person}{Konstantinos Makantasis}, \bibinfo{person}{Antonios Liapis}, {and} \bibinfo{person}{Georgios~N Yannakakis}.} \bibinfo{year}{2022}\natexlab{a}.
\newblock \showarticletitle{RankNEAT: outperforming stochastic gradient search in preference learning tasks}. In \bibinfo{booktitle}{\emph{Proceedings of the Genetic and Evolutionary Computation Conference}}. \bibinfo{pages}{1084--1092}.
\newblock


\bibitem[Pinitas et~al\mbox{.}(2022b)]%
        {pinitas2022supervised}
\bibfield{author}{\bibinfo{person}{Kosmas Pinitas}, \bibinfo{person}{Konstantinos Makantasis}, \bibinfo{person}{Antonios Liapis}, {and} \bibinfo{person}{Georgios~N Yannakakis}.} \bibinfo{year}{2022}\natexlab{b}.
\newblock \showarticletitle{Supervised contrastive learning for affect modelling}. In \bibinfo{booktitle}{\emph{Proceedings of the International Conference on Multimodal Interaction}}. \bibinfo{pages}{531--539}.
\newblock


\bibitem[Pinitas et~al\mbox{.}(2025)]%
        {pinitas2025across}
\bibfield{author}{\bibinfo{person}{Kosmas Pinitas}, \bibinfo{person}{Konstantinos Makantasis}, {and} \bibinfo{person}{Georgios~N Yannakakis}.} \bibinfo{year}{2025}\natexlab{}.
\newblock \showarticletitle{Across-game engagement modelling via few-shot learning}. In \bibinfo{booktitle}{\emph{European Conference on Computer Vision}}. Springer, \bibinfo{pages}{390--406}.
\newblock


\bibitem[Pinitas et~al\mbox{.}({[n.\,d.]})]%
        {pinitas2024varying}
\bibfield{author}{\bibinfo{person}{Kosmas Pinitas}, \bibinfo{person}{Nemanja Rasajski}, \bibinfo{person}{Matthew Barthet}, \bibinfo{person}{Maria Kaselimi}, \bibinfo{person}{Konstantinos Makantasis}, \bibinfo{person}{Antonios Liapis}, {and} \bibinfo{person}{Georgios~N Yannakakis}.} \bibinfo{year}{[n.\,d.]}\natexlab{}.
\newblock \showarticletitle{Varying the context to advance affect modelling: A study on game engagement prediction}.
\newblock


\bibitem[Pinitas et~al\mbox{.}(2024)]%
        {pinitas2024silhouette}
\bibfield{author}{\bibinfo{person}{Kosmas Pinitas}, \bibinfo{person}{Nemanja Rasajski}, \bibinfo{person}{Konstantinos Makantasis}, {and} \bibinfo{person}{Georgios~N Yannakakis}.} \bibinfo{year}{2024}\natexlab{}.
\newblock \showarticletitle{Silhouette Distance Loss for Learning Few-Shot Contrastive Representations}.
\newblock \bibinfo{journal}{\emph{Proceedings of Machine Learning Research}}  \bibinfo{volume}{1} (\bibinfo{year}{2024}), \bibinfo{pages}{18}.
\newblock


\bibitem[Pinitas et~al\mbox{.}(2023)]%
        {pinitas2023predicting}
\bibfield{author}{\bibinfo{person}{Kosmas Pinitas}, \bibinfo{person}{David Renaudie}, \bibinfo{person}{Mike Thomsen}, \bibinfo{person}{Matthew Barthet}, \bibinfo{person}{Konstantinos Makantasis}, \bibinfo{person}{Antonios Liapis}, {and} \bibinfo{person}{Georgios~N Yannakakis}.} \bibinfo{year}{2023}\natexlab{}.
\newblock \showarticletitle{Predicting Player Engagement in Tom Clancy's The Division 2: A Multimodal Approach via Pixels and Gamepad Actions}. In \bibinfo{booktitle}{\emph{Proceedings of the 25th International Conference on Multimodal Interaction}}. \bibinfo{pages}{488--497}.
\newblock


\bibitem[Ringeval et~al\mbox{.}(2013)]%
        {ringeval2013introducing}
\bibfield{author}{\bibinfo{person}{Fabien Ringeval}, \bibinfo{person}{Andreas Sonderegger}, \bibinfo{person}{Juergen Sauer}, {and} \bibinfo{person}{Denis Lalanne}.} \bibinfo{year}{2013}\natexlab{}.
\newblock \showarticletitle{Introducing the RECOLA multimodal corpus of remote collaborative and affective interactions}. In \bibinfo{booktitle}{\emph{Proc. of the IEEE Int. conf. and workshops on automatic face and gesture recognition}}.
\newblock


\bibitem[S{\'a}nchez-Reolid et~al\mbox{.}(2022)]%
        {sanchez2022machine}
\bibfield{author}{\bibinfo{person}{Roberto S{\'a}nchez-Reolid}, \bibinfo{person}{Francisco L{\'o}pez de~la Rosa}, \bibinfo{person}{Daniel S{\'a}nchez-Reolid}, \bibinfo{person}{Mar{\'\i}a~T L{\'o}pez}, {and} \bibinfo{person}{Antonio Fern{\'a}ndez-Caballero}.} \bibinfo{year}{2022}\natexlab{}.
\newblock \showarticletitle{Machine learning techniques for arousal classification from electrodermal activity: A systematic review}.
\newblock \bibinfo{journal}{\emph{Sensors}} \bibinfo{volume}{22}, \bibinfo{number}{22} (\bibinfo{year}{2022}), \bibinfo{pages}{8886}.
\newblock


\bibitem[Sharmanska et~al\mbox{.}(2013)]%
        {sharmanska2013learning}
\bibfield{author}{\bibinfo{person}{Viktoriia Sharmanska}, \bibinfo{person}{Novi Quadrianto}, {and} \bibinfo{person}{Christoph~H Lampert}.} \bibinfo{year}{2013}\natexlab{}.
\newblock \showarticletitle{Learning to rank using privileged information}. In \bibinfo{booktitle}{\emph{Proc. of the IEEE Int. Conf. on computer vision}}. \bibinfo{pages}{825--832}.
\newblock


\bibitem[Souter et~al\mbox{.}(2023)]%
        {souter2023valence}
\bibfield{author}{\bibinfo{person}{Nicholas~E Souter}, \bibinfo{person}{Ariyana Reddy}, \bibinfo{person}{Jake Walker}, \bibinfo{person}{Juli{\'a}n Marino~D{\'a}volos}, {and} \bibinfo{person}{Elizabeth Jefferies}.} \bibinfo{year}{2023}\natexlab{}.
\newblock \showarticletitle{How do valence and meaning interact? The contribution of semantic control}.
\newblock \bibinfo{journal}{\emph{Journal of Neuropsychology}} \bibinfo{volume}{17}, \bibinfo{number}{3} (\bibinfo{year}{2023}), \bibinfo{pages}{521--539}.
\newblock


\bibitem[Sun et~al\mbox{.}(2020)]%
        {sun2020dynamic}
\bibfield{author}{\bibinfo{person}{Bo Sun}, \bibinfo{person}{Siming Cao}, \bibinfo{person}{Dongliang Li}, \bibinfo{person}{Jun He}, {and} \bibinfo{person}{Lejun Yu}.} \bibinfo{year}{2020}\natexlab{}.
\newblock \showarticletitle{Dynamic micro-expression recognition using knowledge distillation}.
\newblock \bibinfo{journal}{\emph{IEEE Transactions on Affective Computing}} \bibinfo{volume}{13}, \bibinfo{number}{2} (\bibinfo{year}{2020}), \bibinfo{pages}{1037--1043}.
\newblock


\bibitem[Tammewar et~al\mbox{.}(2019)]%
        {tammewar2019modeling}
\bibfield{author}{\bibinfo{person}{Aniruddha Tammewar}, \bibinfo{person}{Alessandra Cervone}, \bibinfo{person}{Eva-Maria Messner}, {and} \bibinfo{person}{Giuseppe Riccardi}.} \bibinfo{year}{2019}\natexlab{}.
\newblock \showarticletitle{Modeling user context for valence prediction from narratives}.
\newblock \bibinfo{journal}{\emph{arXiv preprint arXiv:1905.05701}} (\bibinfo{year}{2019}).
\newblock


\bibitem[Toisoul et~al\mbox{.}(2021)]%
        {toisoul_estimation_2021}
\bibfield{author}{\bibinfo{person}{Antoine Toisoul}, \bibinfo{person}{Jean Kossaifi}, \bibinfo{person}{Adrian Bulat}, \bibinfo{person}{Georgios Tzimiropoulos}, {and} \bibinfo{person}{Maja Pantic}.} \bibinfo{year}{2021}\natexlab{}.
\newblock \showarticletitle{Estimation of continuous valence and arousal levels from faces in naturalistic conditions}.
\newblock \bibinfo{journal}{\emph{Nature Machine Intelligence}} \bibinfo{volume}{3}, \bibinfo{number}{1} (\bibinfo{date}{Jan.} \bibinfo{year}{2021}), \bibinfo{pages}{42--50}.
\newblock
\href{https://doi.org/10.1038/s42256-020-00280-0}{doi:\nolinkurl{10.1038/s42256-020-00280-0}}


\bibitem[Trigeorgis et~al\mbox{.}(2016)]%
        {trigeorgis_adieu_2016}
\bibfield{author}{\bibinfo{person}{George Trigeorgis}, \bibinfo{person}{Fabien Ringeval}, \bibinfo{person}{Raymond Brueckner}, \bibinfo{person}{Erik Marchi}, \bibinfo{person}{Mihalis~A. Nicolaou}, \bibinfo{person}{Björn Schuller}, {and} \bibinfo{person}{Stefanos Zafeiriou}.} \bibinfo{year}{2016}\natexlab{}.
\newblock \showarticletitle{Adieu features? {End}-to-end speech emotion recognition using a deep convolutional recurrent network}. In \bibinfo{booktitle}{\emph{Proc. of the Int. Conf. on {Acoustics}, {Speech} and {Signal} {Processing} ({ICASSP})}}. \bibinfo{pages}{5200--5204}.
\newblock
\href{https://doi.org/10.1109/ICASSP.2016.7472669}{doi:\nolinkurl{10.1109/ICASSP.2016.7472669}}


\bibitem[Tzirakis et~al\mbox{.}(2021)]%
        {tzirakis_end--end_2021}
\bibfield{author}{\bibinfo{person}{Panagiotis Tzirakis}, \bibinfo{person}{Jiaxin Chen}, \bibinfo{person}{Stefanos Zafeiriou}, {and} \bibinfo{person}{Björn Schuller}.} \bibinfo{year}{2021}\natexlab{}.
\newblock \showarticletitle{End-to-end multimodal affect recognition in real-world environments}.
\newblock \bibinfo{journal}{\emph{Information Fusion}}  \bibinfo{volume}{68} (\bibinfo{date}{April} \bibinfo{year}{2021}), \bibinfo{pages}{46--53}.
\newblock
\href{https://doi.org/10.1016/j.inffus.2020.10.011}{doi:\nolinkurl{10.1016/j.inffus.2020.10.011}}


\bibitem[Vapnik and Izmailov(2015)]%
        {vapnik2015learning}
\bibfield{author}{\bibinfo{person}{Vladimir Vapnik} {and} \bibinfo{person}{Rauf Izmailov}.} \bibinfo{year}{2015}\natexlab{}.
\newblock \showarticletitle{Learning using privileged information: similarity control and knowledge transfer.}
\newblock \bibinfo{journal}{\emph{Journal of Machine Learning Research}} \bibinfo{volume}{16}, \bibinfo{number}{1} (\bibinfo{year}{2015}), \bibinfo{pages}{2023--2049}.
\newblock


\bibitem[Vapnik and Izmailov(2017)]%
        {vapnik2017knowledge}
\bibfield{author}{\bibinfo{person}{Vladimir Vapnik} {and} \bibinfo{person}{Rauf Izmailov}.} \bibinfo{year}{2017}\natexlab{}.
\newblock \showarticletitle{Knowledge transfer in SVM and neural networks}.
\newblock \bibinfo{journal}{\emph{Annals of Mathematics and Artificial Intelligence}} \bibinfo{volume}{81}, \bibinfo{number}{1} (\bibinfo{year}{2017}), \bibinfo{pages}{3--19}.
\newblock


\bibitem[Vapnik and Vashist(2009)]%
        {vapnik2009new}
\bibfield{author}{\bibinfo{person}{Vladimir Vapnik} {and} \bibinfo{person}{Akshay Vashist}.} \bibinfo{year}{2009}\natexlab{}.
\newblock \showarticletitle{A new learning paradigm: Learning using privileged information}.
\newblock \bibinfo{journal}{\emph{Neural networks}} \bibinfo{volume}{22}, \bibinfo{number}{5-6} (\bibinfo{year}{2009}), \bibinfo{pages}{544--557}.
\newblock


\bibitem[Wang et~al\mbox{.}(2020)]%
        {wang2020generalizing}
\bibfield{author}{\bibinfo{person}{Yaqing Wang}, \bibinfo{person}{Quanming Yao}, \bibinfo{person}{James~T Kwok}, {and} \bibinfo{person}{Lionel~M Ni}.} \bibinfo{year}{2020}\natexlab{}.
\newblock \showarticletitle{Generalizing from a few examples: A survey on few-shot learning}.
\newblock \bibinfo{journal}{\emph{ACM computing surveys (csur)}} \bibinfo{volume}{53}, \bibinfo{number}{3} (\bibinfo{year}{2020}), \bibinfo{pages}{1--34}.
\newblock


\bibitem[Wieser and Brosch(2012)]%
        {wieser2012faces}
\bibfield{author}{\bibinfo{person}{Matthias~J Wieser} {and} \bibinfo{person}{Tobias Brosch}.} \bibinfo{year}{2012}\natexlab{}.
\newblock \showarticletitle{Faces in context: A review and systematization of contextual influences on affective face processing}.
\newblock \bibinfo{journal}{\emph{Frontiers in psychology}}  \bibinfo{volume}{3} (\bibinfo{year}{2012}), \bibinfo{pages}{471}.
\newblock


\bibitem[Yang et~al\mbox{.}(2023)]%
        {yang2023cluster}
\bibfield{author}{\bibinfo{person}{Kailai Yang}, \bibinfo{person}{Tianlin Zhang}, \bibinfo{person}{Hassan Alhuzali}, {and} \bibinfo{person}{Sophia Ananiadou}.} \bibinfo{year}{2023}\natexlab{}.
\newblock \showarticletitle{Cluster-level contrastive learning for emotion recognition in conversations}.
\newblock \bibinfo{journal}{\emph{IEEE Transactions on Affective Computing}} (\bibinfo{year}{2023}).
\newblock


\bibitem[Yannakakis et~al\mbox{.}(2018)]%
        {yannakakis2018ordinal}
\bibfield{author}{\bibinfo{person}{Georgios~N Yannakakis}, \bibinfo{person}{Roddy Cowie}, {and} \bibinfo{person}{Carlos Busso}.} \bibinfo{year}{2018}\natexlab{}.
\newblock \showarticletitle{The ordinal nature of emotions: An emerging approach}.
\newblock \bibinfo{journal}{\emph{IEEE Transactions on Affective Computing}} \bibinfo{volume}{12}, \bibinfo{number}{1} (\bibinfo{year}{2018}), \bibinfo{pages}{16--35}.
\newblock


\bibitem[Yoon et~al\mbox{.}(2005)]%
        {yoon2005feature}
\bibfield{author}{\bibinfo{person}{Hyunjin Yoon}, \bibinfo{person}{Kiyoung Yang}, {and} \bibinfo{person}{Cyrus Shahabi}.} \bibinfo{year}{2005}\natexlab{}.
\newblock \showarticletitle{Feature subset selection and feature ranking for multivariate time series}.
\newblock \bibinfo{journal}{\emph{IEEE transactions on knowledge and data engineering}} \bibinfo{volume}{17}, \bibinfo{number}{9} (\bibinfo{year}{2005}), \bibinfo{pages}{1186--1198}.
\newblock


\bibitem[Zhang and Etemad(2021)]%
        {zhang2021distilling}
\bibfield{author}{\bibinfo{person}{Guangyi Zhang} {and} \bibinfo{person}{Ali Etemad}.} \bibinfo{year}{2021}\natexlab{}.
\newblock \showarticletitle{Distilling EEG representations via capsules for affective computing}.
\newblock \bibinfo{journal}{\emph{arXiv preprint arXiv:2105.00104}} (\bibinfo{year}{2021}).
\newblock


\end{thebibliography}

\end{document}